\documentclass{article}

\usepackage{microtype}
\usepackage{graphicx}
\usepackage{subcaption}
\usepackage{booktabs} 

\usepackage{hyperref}

\usepackage[preprint]{icml2026}

\usepackage{amsmath}
\usepackage{amssymb}
\usepackage{mathtools}
\usepackage{amsthm}

\usepackage[capitalize,noabbrev]{cleveref}

\theoremstyle{plain}

\theoremstyle{definition}

\theoremstyle{remark}

\usepackage[textsize=tiny]{todonotes}

\usepackage[utf8]{inputenc}
\usepackage[T1]{fontenc}
\usepackage{algorithm}
\usepackage{algorithmic}
\usepackage{graphicx}
\usepackage{booktabs}
\usepackage{multirow}
\usepackage{amsmath}
\usepackage{amsfonts}
\usepackage{enumitem}
\usepackage{url} 
\usepackage{xspace}
\usepackage[table]{xcolor}
\usepackage{hyperref}
\usepackage{epigraph}
\usepackage{amsthm}
\usepackage{amssymb}
\usepackage{arydshln}
\usepackage{bbm}
\usepackage{wrapfig}

\definecolor{langlightblue}{rgb}{0.3, 0.65, 1}
\definecolor{langblue}{rgb}{0, 0.4, 0.8}
\definecolor{langmildblue}{rgb}{0.0, 0.45, 0.73}
\definecolor{langdarkblue}{rgb}{0.0, 0.0, 0.61}
\definecolor{langred}{rgb}{0.81, 0.09, 0.13}
\definecolor{langlightgreen}{rgb}{0.80, 0.97, 0.85}
\definecolor{langgreen}{rgb}{0.18, 0.55, 0.34}
\definecolor{langdarkgreen}{rgb}{0.0, 0.45, 0.38}
\definecolor{bingpink}{rgb}{1.0, 0.41, 0.71}
\hypersetup{
    colorlinks=true,
    linkcolor=langdarkgreen,
    citecolor=langgreen,
    urlcolor=langdarkgreen,
}
\newcommand{\gain}[1]{\textsubscript{\textbf{\textit{\textcolor{langdarkgreen}{#1}}}}}
\newcommand{\loss}[1]{\textsubscript{\textbf{\textit{\textcolor{langred}{#1}}}}}

\makeatletter
\AtBeginDocument{%
  \let\oldref\ref
  \renewcommand{\ref}[1]{%
    \hyperref[{#1}]{\underline{\oldref{#1}}}%
  }%
}
\makeatother

\usepackage{titletoc}

\newcommand\method{\textit{EHR-RAG}\xspace}

\icmltitlerunning{EHR-RAG}

\begin{document}

\twocolumn[
  \icmltitle{EHR-RAG: Bridging Long-Horizon Structured Electronic Health Records and Large Language Models via Enhanced Retrieval-Augmented Generation}



  \icmlsetsymbol{equal}{*}

    \begin{icmlauthorlist}
    \icmlauthor{Lang Cao}{uiuc}
    \icmlauthor{Qingyu Chen}{yale}
    \icmlauthor{Yue Guo}{uiuc}
    \end{icmlauthorlist}
    
    
    \icmlaffiliation{uiuc}{University of Illinois Urbana-Champaign, United States}
    \icmlaffiliation{yale}{Yale University, United States}
    
    \icmlcorrespondingauthor{Lang Cao}{langcao2@illinois.edu}
    \icmlcorrespondingauthor{Yue Guo}{yueg@illinois.edu}

  \icmlkeywords{Machine Learning, ICML}

  \vskip 0.3in
]



\printAffiliationsAndNotice{}  

\begin{abstract}
Electronic Health Records (EHRs) provide rich longitudinal clinical evidence that is central to medical decision-making, motivating the use of retrieval-augmented generation (RAG) to ground large language model (LLM) predictions. However, long-horizon EHRs often exceed LLM context limits, and existing approaches commonly rely on truncation or vanilla retrieval strategies that discard clinically relevant events and temporal dependencies. To address these challenges, we propose \method, a retrieval-augmented framework designed for accurate interpretation of long-horizon structured EHR data. \method introduces three components tailored to longitudinal clinical prediction tasks: \textit{Event- and Time-Aware Hybrid EHR Retrieval} to preserve clinical structure and temporal dynamics, \textit{Adaptive Iterative Retrieval} to progressively refine queries in order to expand broad evidence coverage, and \textit{Dual-Path Evidence Retrieval and Reasoning} to jointly retrieves and reasons over both factual and counterfactual evidence. Experiments across four long-horizon EHR prediction tasks show that \method consistently outperforms the strongest LLM-based baselines, achieving an average Macro-F1 improvement of 10.76\%. Overall, our work highlights the potential of retrieval-augmented LLMs to advance clinical prediction on structured EHR data in practice.

\end{abstract}


\section{Introduction}
Electronic Health Records (EHRs) are digital longitudinal patient records composed of structured clinical events such as diagnoses, medications, laboratory results, and procedures. Unlike free-text clinical notes, structured EHR data are systematically collected, standardized across care settings, and less subjective in description, making them a reliable and widely available foundation for clinical decision making \cite{rosenbloom2011data, ebbers2022impact}. Beyond individual events, structured EHRs encode rich temporal and categorical information, including event types, ordering, recurrence patterns, and long-term clinical trajectories, that underpins clinical decision-making across the full course of clinical care \cite{menachemi2011benefits, cowie2017electronic}. For many non-acute or chronic conditions, patients accumulate long-horizon EHRs spanning years of irregular visits and thousands of heterogeneous events. These trajectories are common in real-world healthcare but difficulty to interpret, even for clinicians, and  recent studies have highlighted both their modeling challenges and clinical importance \cite{loh2025limitations, wornow2023ehrshot, fries2025longitudinalEHR}. Effective clinical prediction in this setting requires reasoning over long temporal spans while preserving event-type structure and clinical dependencies.

Traditional machine learning (ML) methods have been widely studied for EHR-based prediction tasks, including length-of-stay estimation \cite{cai2016real, levin2021machine}, readmission prediction \cite{ashfaq2019readmission, xiao2018readmission}, and medication recommendation \cite{bhoi2021personalizing, shang2019gamenet}. More recently, large language models (LLMs) \cite{achiam2023gpt, touvron2023llama} have shown promise in clinical reasoning due to their strong generalization capabilities to new task and their capability to perform multi-step reasoning to provide rationale instead of only final answer to make it more trustworthy for clinician's decision making \cite{bedi2025testing, jiang2023graphcare, jiang2024reasoning, kojima2022large}. However, LLMs cannot natively process long-horizon structured EHRs: their fixed context windows are quickly exceeded by multi-visit patient histories, and naive event serialization obscures temporal structure and discards clinically relevant context.

Existing approaches address this limitation primarily through heuristic truncation, such as retaining only recent events \cite{lin2025training} or selecting events associated with high-frequency codes \cite{liao2025ehr}. While retrieval-augmented generation (RAG) \cite{gao2023retrieval} offers a mechanism to selectively retrieve relevant events, existing RAG-based methods typically retrieve isolated subsets of the patient history primarily to satisfy context-length constraints \cite{zhu2024realm, kruse2025large, zhu2024emerge}. As a result, vanilla RAG in clinical settings often suffers from low retrieval quality, incomplete evidence coverage, and reduced robustness when reasoning over long-horizon EHRs \cite{li2025muisqa,perccin2025investigating}. Despite the prevalence and importance of such data, to the best of our knowledge, no existing RAG framework is explicitly designed to handle their longitudinal, multi-event nature.

In this work, we study clinical prediction with long-horizon structured EHR data and propose \textbf{\method}, an enhanced retrieval-augmented framework tailored for this setting. 
\method is built around three components designed to improve retrieval \textbf{\textit{quality}}, \textbf{\textit{completeness}}, and \textbf{\textit{robustness}} under strict context constraints:
(a) \textit{\textbf{E}vent- and \textbf{T}ime-Aware \textbf{H}ybrid \textbf{E}HR \textbf{R}etrieval} (\textbf{\textit{ETHER}}), which preserves event-type structure and temporal dynamics during retrieval;
(b) \textit{\textbf{A}daptive \textbf{I}terative \textbf{R}etrieval} (\textbf{\textit{AIR}}), which progressively refines retrieval queries to expand evidence coverage while respecting context limits; and
(c) \textit{\textbf{D}ual-Path \textbf{E}vidence \textbf{R}etrieval and \textbf{R}easoning} (\textbf{\textit{DER}}), which jointly retrieves and reasons over both factual patient history and counterfactual evidence to improve robustness. Together, these designs enable reliable long-horizon clinical reasoning with LLMs. We evaluate \method against a range of baseline methods on a long-horizon EHR benchmark. Experimental results demonstrate consistent Macro-F1 improvements across all four clinical prediction tasks, with gains of 3.63\% on \textit{Long Length of Stay}, 11.28\% on \textit{30-day Readmission}, 16.46\% on \textit{Acute Myocardial Infarction}, and 7.66\% on \textit{Anemia} relative to the strongest LLM-based baselines. Overall, \method achieves an average Macro-F1 improvement of 10.76\%.

In summary, we make the following contributions:
\begin{itemize}[leftmargin=*, itemsep=0pt, labelsep=5pt, topsep=0pt]
    \item We identify and analyze the limitations of existing truncation-based and vanilla RAG approaches on long-horizon structured EHR data, highlighting challenges in evidence completeness and temporal reasoning for reliable clinical prediction.
    \item We propose \method, a retrieval-augmented framework explicitly designed for long-horizon EHR reasoning, integrating event- and time-aware retrieval, adaptive iterative retrieval, dual-path evidence retrieval. 
    \item We conduct extensive experiments demonstrating consistent improvements over strong baselines across multiple long-horizon EHR prediction tasks.
\end{itemize}


\section{Related Work}
\begin{figure*}[htbp]
    \centering
    \includegraphics[width=\textwidth]{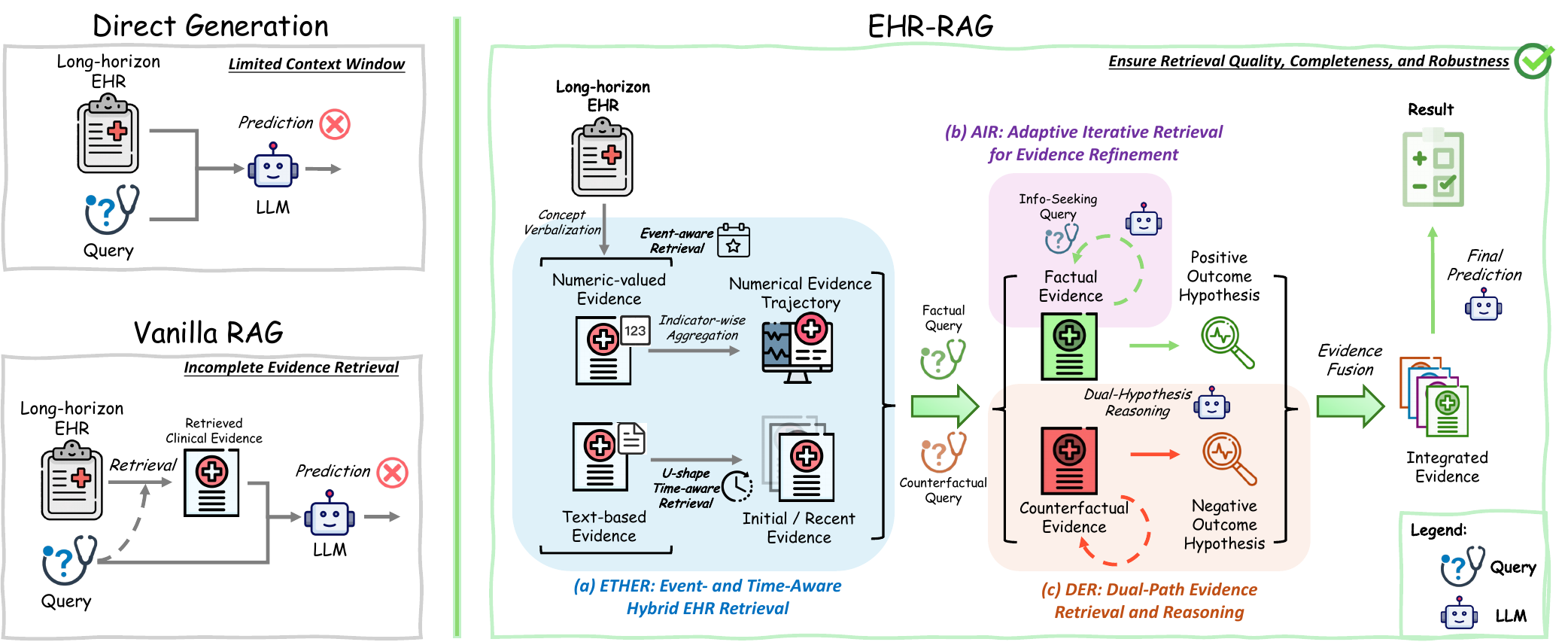}
    \caption{Overview of the \textbf{\method} framework for long-horizon clinical prediction. Compared with \textit{Direct Generation} and \textit{Vanilla RAG}, our framework explicitly addresses context truncation and incomplete retrieval. It integrates \textbf{\textit{(a) Event- and Time-aware Hybrid Retrieval (ETHER)}}, \textbf{\textit{(b) Adaptive Iterative Retrieval (AIR)}}, and \textbf{\textit{(c) Dual-Path Factual and Counterfactual Reasoning (DER)}}, ensuring retrieval quality, robustness, and completeness for reliable clinical decision-making.}
    \label{fig:framework}
\end{figure*}

\noindent\textbf{Clinical Predictive Models.}
Structured EHR data has supported the development of a wide range of machine learning models for different clinical prediction tasks \cite{cai2016real, levin2021machine, ashfaq2019readmission, bhoi2021personalizing}. Models such as RETAIN \cite{choi2016retain}, GRAM \cite{choi2017gram}, and KerPrint \cite{yang2023kerprint} are specifically designed to capture complex temporal and hierarchical patterns within structured EHRs and have shown strong performance across multiple predictive settings. In parallel, recent efforts have focused on scaling clinical prediction through large foundation models, including CLMBR-T-Base \cite{wornow2023ehrshot}, Virchow \cite{vorontsov2024foundation}, and MIRA \cite{li2025mira}. However, traditional predictive models—ranging from task-specific architectures to supervised foundation model training—remain inflexible, requiring labeled data for every downstream task and often failing to generalize beyond the distribution on which they were trained. These limitations are especially problematic in the dynamic and heterogeneous healthcare environment.

\noindent\textbf{LLM-based Clinical Prediction.}
Recently, the paradigm of clinical prediction has begun to shift toward the use of LLMs \cite{achiam2023gpt}. LLMs offer greater adaptability and generalization than traditional supervised models, with the capacity to interpret diverse medical information and support more versatile clinical decision-making. Several studies apply LLMs directly to clinical prediction tasks \cite{lovon2024revisiting, zhu2024prompting, chen2024clinicalbench, kruse2025large}, while others employ LLMs as encoders for structured EHR data \cite{hegselmann2025large}. Additional efforts explore training or adapting LLMs for specific downstream healthcare tasks \cite{lin2025training, jiang2023graphcare, jiang2024reasoning, yang2022large}, and recent work introduces LLM-driven agents to assist with clinical reasoning and EHR interaction \cite{cui2025llms, shi2024ehragent}. Despite these advances, few studies have addressed long-horizon EHR prediction, leaving substantial challenges in enabling LLMs to interpret complex, multi-visit longitudinal patient histories. Our work extends this line of research by developing an LLM-based approach specifically tailored for long-horizon EHR prediction.

\noindent\textbf{Retrieval-Augmented Generation with LLM.}
RAG has emerged as an effective technique for enhancing LLMs with external knowledge beyond their limited context window \cite{gao2023retrieval}. A variety of retrieval strategies have been proposed to help LLMs accurately access additional relevant information, leading to improved performance across diverse tasks and settings \cite{asai2024self, jiang2025deepretrieval, jiang2025ras}. In the healthcare domain, where clinical information is often overwhelming in volume and complexity, RAG has become increasingly popular for grounding LLM outputs in accurate task-relevant evidence \cite{neha2025retrieval, kim2025rethinking}. Some studies use RAG to inject medical knowledge and improve healthcare applications \cite{soman2024biomedical, niu2024ehr, cao2024autord}, while others employ RAG to mitigate long-context limitations. For example, REALM \cite{zhu2024realm} enhances multimodal EHR analysis using LLMs, and EMERGE \cite{zhu2024emerge} integrates RAG into multimodal EHR predictive modeling. Additional work has investigated RAG-based methods for clinical trial and patient matching \cite{tramontini2025investigating}.However, most of these approaches focus on unstructured EHR data such as clinical notes, rather than structured tabular EHR, which exhibits distinct temporal dynamics and event-type structure. Moreover, vanilla RAG pipelines often suffer from incomplete evidence retrieval in clinical settings. Our work aims to fill this gap by designing a RAG framework specifically tailored for long-horizon structured EHR data.

\section{Methodology}

Figure~\ref{fig:framework} illustrates \method, a retrieval-augmented framework designed to enable LLMs to reason over long-horizon structured EHR data under strict context constraints. The framework comprises three core components: \textit{Event- and Time-Aware Hybrid EHR Retrieval (ETHER)}, \textit{Adaptive Iterative Retrieval (AIR)}, and \textit{Dual-Path Evidence Retrieval and Reasoning (DER)}.

\subsection{Task Formulation}
\label{sec:task_formulation}
Given a patient $p$, we assume access to a longitudinal structured EHR 
$\mathcal{E}_p = \{ e_i \}_{i=1}^{N}$, where each clinical event is represented as $e_i = (c_i, v_i, \tau_i)$, with $c_i$ denoting a clinical concept (e.g., diagnosis, procedure, laboratory test, or medication),
$v_i$ the associated value (numeric or textual), and $\tau_i$ the timestamp at which the event occurred.
Events are ordered chronologically such that
$\tau_1 \le \tau_2 \le \cdots \le \tau_N$.

At a prediction time $\tau^{\ast}$, the available patient history is
\begin{equation}
\mathcal{E}_p^{\le \tau^{\ast}} = \{ e_i \in \mathcal{E}_p \mid \tau_i \le \tau^{\ast} \}.
\end{equation}
Our goal is to predict a clinical outcome $y \in \mathcal{Y}$ (binary or multiclass) based on this long-horizon EHR context.
Examples include prolonged length of stay, laboratory abnormality severity, or risk of acute clinical events.

We assume access to a pretrained LLM $\mathcal{M}$ capable of clinical reasoning
but limited by a fixed context window, making directly input of
$\mathcal{E}_p^{\le \tau^{\ast}}$ infeasible when $N$ is large.
Therefore, we aim to construct a compact, task-relevant evidence context $\mathcal{C} \subset \mathcal{E}_p^{\le \tau^{\ast}}$ such that $\hat{y} = \mathcal{M}(\mathcal{C})$, is consistent with the ground-truth outcome $y$.

\subsection{Event- and Time-Aware Hybrid EHR Retrieval}

Structured EHR $\mathcal{E}_p^{\le \tau^\ast}$ contain heterogeneous event types, including numeric measurements (e.g., lab tests and vital signs) and textual records (e.g., event descriptions and clinical notes). Treating all clinical events uniformly during retrieval is often ineffective. Dense text embeddings are primarily optimized for capturing semantic similarity and are known to be insensitive to precise numerical values and magnitude differences \cite{wallace2019nlp,zhang2020language}, which limits their effectiveness for reasoning over numeric clinical measurements such as laboratory results and vital signs. Moreover, raw numerical values alone do not fully convey their clinical significance without appropriate temporal context and longitudinal interpretation. Motivated by these limitations, we propose an event- and time-aware hybrid retrieval strategy that explicitly handles numeric events separately while preserving their temporal structure.

\textbf{Indicator-wise aggregation of numeric events.}
All numeric clinical events are first grouped by clinical indicator (i.e., event name), forming a collection of indicator-specific temporal trajectories. Formally, for each indicator $k$, we define its numeric value trajectory as $\mathcal{V}_k = \left\{ \left(v_{k,j}, \tau_{k,j}\right) \;\middle|\; \tau_{k,j} \le \tau^\ast \right\}$, where $v_{k,j}$ denotes the $j$-th observed numeric value of indicator $k$ and $\tau_{k,j}$ is its corresponding timestamp. Collectively, the set $\{\mathcal{V}_k\}$ forms an indicator-level numeric representation of the patient, where each $\mathcal{V}_k$ captures the longitudinal evolution of a single clinical measurement. This representation preserves clinically meaningful temporal dynamics while enabling efficient and semantically informed retrieval over numeric EHR data.

\textbf{Coarse-to-fine indicator selection.}
Given a task-specific query $q_{\text{task}}$, we first perform coarse-grained retrieval over all indicator-level summaries $\{\mathcal{V}_k\}$ using dense similarity search, where embeddings are computed from indicator names. This step yields a candidate indicator set $\mathcal{K}_{\text{coarse}} = \{k_1, \ldots, k_{N_{\text{coarse}}}\}$, where $N_{\text{coarse}}$ denotes the number of indicators retained after coarse retrieval. We then apply LLM-based reranking to select a smaller, task-relevant subset $\mathcal{K}_{\text{fine}} \subset \mathcal{K}_{\text{coarse}}$ such that $|\mathcal{K}_{\text{fine}}| = N_{\text{fine}}$. For each selected indicator $k \in \mathcal{K}_{\text{fine}}$, we retain the $N_{\text{recent}}$ most recent measurements occurring prior to the prediction time $\tau^\ast$ to ensure temporal relevance, forming an indicator-specific numeric evidence trajectory $\mathcal{E}_k = \{(v_{k,j}, \tau_{k,j}) \in \mathcal{V}_k \mid \tau_{k,j} \le \tau^\ast\}$. Collectively, these indicator-level evidence trajectories constitute the final numeric evidence collection $\mathcal{E}_{\text{num}} = \{\mathcal{E}_k \mid k \in \mathcal{K}_{\text{fine}}\}$.

\textbf{U-shape time-aware retrieval for textual events.}
Non-numeric clinical events (e.g., diagnoses, procedures, and clinical notes) are serialized at the event level and segmented into overlapping temporal chunks, with each chunk embedded and indexed in a vector store. Given a task-specific query $q_{\text{task}}$ and prediction time $\tau^\ast$, we first retrieve a candidate pool of $K_{\text{cand}}$ textual chunks using dense semantic similarity search, and then jointly consider semantic relevance and temporal proximity to select textual evidence.

Prior work has explored incorporating temporal information to improve retrieval quality, for example by introducing time-aware relevance weighting or decay functions~\cite{cao2024pilot,abdallah2025extending}. However, in clinical settings, temporal importance is not strictly monotonic: while recent events are often critical, early events corresponding to disease onset or initial admission can be equally informative. This observation motivates a U-shaped time-aware retrieval strategy that explicitly emphasizes both recent and early clinical evidence.

For each candidate textual chunk $c$ with timestamp $\tau_c$, we compute a semantic similarity score $s_{\text{sem}}(q_{\text{task}}, c)$ using dense embeddings, together with a U-shaped temporal relevance score. Let $\tau_{\text{first}}$ denote the earliest timestamp in the patient record. The temporal relevance score is defined as
\begin{equation}
s_{\text{time}}(\tau_c)
=
\max \left(
\exp\!\left(-\frac{\tau^\ast - \tau_c}{\tau_{\text{recent}}}\right),
\exp\!\left(-\frac{\tau_c - \tau_{\text{first}}}{\tau_{\text{early}}}\right)
\right).
\end{equation}

This U-shaped formulation assigns higher importance to events occurring close to the prediction time $\tau^\ast$ or near the beginning of the patient trajectory, while downweighting mid-history events that are less informative for the current prediction. The final retrieval score for each textual chunk is computed as a convex combination of semantic relevance and temporal importance,
\begin{equation}
s(c)
=
\alpha \, s_{\text{sem}}(q_{\text{task}}, c)
+
(1 - \alpha) \, s_{\text{time}}(\tau_c),
\end{equation}
where $\alpha \in [0,1]$ controls the trade-off between semantic similarity and temporal relevance. Candidate chunks are ranked by $s(c)$, and the top-$K_{\text{final}}$ chunks are selected and temporally ordered to form the final textual evidence set $\mathcal{E}_{\text{text}} = \{ c_1, \ldots, c_{K_{\text{final}}} \}$.

By combining indicator-level numeric evidence $\mathcal{E}_{\text{num}}$ with U-shaped time-aware textual evidence $\mathcal{E}_{\text{text}}$, \textit{ETHER} ensures broad semantic coverage while preserving event-type distinctions and clinically meaningful temporal structure under a compact evidence budget.

\subsection{Adaptive Iterative Retrieval}
Single-pass retrieval is often insufficient for long-horizon EHRs, where relevant clinical evidence may be temporally dispersed or only indirectly related to the initial query. To address this limitation, we progressively expands evidence coverage in a controlled and targeted manner.

Starting from an initial query $q^{(0)} = q_{\text{task}}$, we retrieve an initial evidence set $\mathcal{E}^{(0)}$. At each iteration $t$, the LLM assesses whether the current evidence set $\mathcal{E}^{(t)}$ is sufficient to answer the clinical prediction task. If so, the retrieval process terminates. Otherwise, the LLM generates a refined query
\begin{equation}
q^{(t+1)} = \mathcal{M}_{\mathcal{R}}\!\left(q^{(t)}, \mathcal{E}^{(t)}\right),
\end{equation}
where $\mathcal{M}_{\mathcal{R}}(\cdot)$ denotes an LLM-based query refinement module that identifies a single missing yet clinically salient aspect of the current evidence. Each refined query is explicitly constrained to be concise, focused on one clinical dimension, and non-redundant with previously retrieved information.

Evidence retrieved using the refined query is then merged with the existing context via deduplication and temporal ordering:
\begin{equation}
\mathcal{E}^{(t+1)} = \mathrm{Merge}\!\left(\mathcal{E}^{(t)}, \mathrm{Retrieve}(q^{(t+1)})\right).
\end{equation}

This iterative process continues until the evidence is deemed sufficient or a predefined iteration limit is reached. By incrementally expanding the evidence set in a targeted manner, AIR improves retrieval recall while preventing uncontrolled context growth, which is critical for effective long-horizon clinical reasoning.

\subsection{Dual-Path Evidence Retrieval and Reasoning}
Clinical prediction often involves competing hypotheses~\citep{sox2024medical,elstein1978medical}, and reasoning along a single evidence pathway can lead to biased or overconfident conclusions~\cite{graber2005diagnostic,croskerry2003importance}. To improve robustness, we propose a dual-path evidence retrieval and reasoning strategy.

Specifically, we construct two complementary retrieval queries: a \textbf{factual (positive) query} $q^{+}$ that seeks evidence supporting the presence of the target outcome, and a \textbf{counterfactual (negative) query} $q^{-}$ that seeks evidence supporting its absence. Each query is processed independently using the adaptive iterative retrieval mechanism, yielding two textual evidence sets, $\mathcal{E}^{+}_{\text{text}}$ and $\mathcal{E}^{-}_{\text{text}}$. For each path, the retrieved textual evidence is combined with the shared numeric evidence $\mathcal{E}_{\text{num}}$, and the LLM is prompted to form an explicit outcome hypothesis:
\begin{equation}
h^{+} = \mathcal{M}\!\left(\mathcal{E}^{+}_{\text{text}} \cup \mathcal{E}_{\text{num}}\right), \quad
h^{-} = \mathcal{M}\!\left(\mathcal{E}^{-}_{\text{text}} \cup \mathcal{E}_{\text{num}}\right),
\end{equation}
where $h^{+}$ and $h^{-}$ denote the positive and negative outcome hypotheses, respectively. The corresponding evidence sets are then fused into a unified evidence context:
\begin{equation}
\mathcal{E}^{\text{fuse}} =
\mathcal{E}^{+}_{\text{text}} \cup
\mathcal{E}^{-}_{\text{text}} \cup
\mathcal{E}_{\text{num}} .
\end{equation}
Finally, the model is explicitly prompted to compare the \textit{strength}, \textit{directness}, and \textit{clinical relevance} of evidence supporting each outcome hypothesis, and to produce a final prediction conditioned on both hypotheses and the fused evidence:
\begin{equation}
\hat{y}
=
\mathcal{M}_{\text{dec}}\!\left(
\mathcal{E}^{\text{fuse}},\,
h^{+},\,
h^{-}
\right),
\end{equation}
where $\mathcal{M}_{\text{dec}}(\cdot)$ denotes the LLM-based decision function that performs comparative hypothesis evaluation and outputs the final clinical prediction.

Overall, this dual-path design enables balanced hypothesis evaluation, mitigates confirmation bias and spurious correlations, and improves the robustness and reliability of long-horizon clinical prediction from structured EHR data.

\begin{table*}[htbp]
\caption{Performance of \method and other LLM-based baselines on \textit{Long Length of Stay} and \textit{30-day Readmission} using \textit{GPT-5}. We report Accuracy (\%), Macro-F1 (\%), and per-class F1 scores (\%). \textbf{Bold} indicates the best performance, and improvements over the strongest baseline are highlighted in \textcolor{langdarkgreen}{green}. \method consistently outperforms other methods on both tasks.}
\label{tab:main_los_readmission}
\centering
\resizebox{0.85\textwidth}{!}{
\begin{tabular}{lcccccccc}
\toprule
\multirow{2}{*}{\textbf{Method}} &
\multicolumn{4}{c}{\textbf{Long Length of Stay}} &
\multicolumn{4}{c}{\textbf{30-day Readmission}} \\
\cmidrule(lr){2-5}\cmidrule(lr){6-9}
& \textbf{Accuracy} & \textbf{Macro F1} & \textbf{F1$_\text{Short}$} & \textbf{F1$_\text{Long}$}
& \textbf{Accuracy} & \textbf{Macro F1} & \textbf{F1$_\text{No Readmit}$} & \textbf{F1$_\text{Readmit}$} \\
\midrule
Direct Generation & 69.41 & 66.52 & 76.36 & 56.67 & 46.56 & 44.37 & 55.41 & 33.33 \\
RAG              & 68.24 & 64.77 & 75.82 & 53.71 & 45.80 & 43.40 & 55.06 & 31.73 \\
Uniform RAG      & 65.10 & 61.50 & 73.27 & 49.72 & 52.67 & 49.63 & 62.65 & 35.42 \\
Rule-based RAG   & 63.53 & 56.58 & 73.95 & 39.22 & 57.63 & 44.08 & 71.61 & 16.54 \\
ReAct RAG        & 65.10 & 60.85 & 73.75 & 47.95 & 48.09 & 45.22 & 57.76 & 32.67 \\
\midrule
\rowcolor{langlightgreen!50}\textbf{EHR-RAG (Ours)} 
& \textbf{74.12\gain{+4.71}} 
& \textbf{70.15\gain{+3.63}} 
& \textbf{81.03\gain{+4.67}} 
& \textbf{59.26\gain{+2.59}} 
& \textbf{71.76\gain{+14.13}} 
& \textbf{60.91\gain{+11.28}} 
& \textbf{81.50\gain{+9.89}} 
& \textbf{40.32\gain{+4.90}} \\
\bottomrule
\end{tabular}
}
\end{table*}
\begin{table*}[htbp]
\caption{Performance of \method and other LLM-based baselines on \textit{Acute Myocardial Infarction} and \textit{Anemia} using \textit{GPT-5}. We report Accuracy (\%), Macro-F1 (\%), and per-class F1 scores (\%). \textbf{Bold} indicates the best performance, and improvements over the strongest baseline are highlighted in \textcolor{langdarkgreen}{green}. \method consistently outperforms other methods on both tasks.}
\label{tab:main_acutemi_anemia}
\centering
\resizebox{\textwidth}{!}{
\begin{tabular}{lcccccccccc}
\toprule
\multirow{2}{*}{\textbf{Method}} &
\multicolumn{4}{c}{\textbf{Acute MI}} &
\multicolumn{6}{c}{\textbf{Anemia}} \\
\cmidrule(lr){2-5}\cmidrule(lr){6-11}
& \textbf{Accuracy} & \textbf{Macro F1} & \textbf{F1$_\text{No MI}$} & \textbf{F1$_\text{MI}$}
& \textbf{Accuracy} & \textbf{Macro F1} & \textbf{F1$_\text{Low}$} & \textbf{F1$_\text{Medium}$} & \textbf{F1$_\text{High}$} & \textbf{F1$_\text{Abnormal}$} \\
\midrule
Direct Generation & 88.38 & 57.97 & 93.72 & 22.22 & 44.57 & 28.38 & 70.74 & 17.30 & 18.48 & 7.02 \\
RAG              & 90.04 & 59.83 & 94.67 & 25.00 & 42.08 & 20.44 & 70.43 & 18.85 & 12.90 & 0.00 \\
Uniform RAG      & 89.21 & 47.15 & 94.30 & 0.00  & 48.42 & 32.39 & 72.65 & 26.97 & 22.68 & 7.27 \\
Rule-based RAG   & 91.29 & 52.06 & 95.42 & 8.70  & 44.34 & 26.21 & 67.61 & 17.33 & 16.30 & 3.57 \\
ReAct RAG        & 89.21 & 56.49 & 94.22 & 18.75 & 46.15 & 31.64 & 69.37 & 27.23 & 22.68 & 7.27 \\
\midrule
\rowcolor{langlightgreen!50}\textbf{EHR-RAG (Ours)} 
& \textbf{92.95}\gain{+1.66}
& \textbf{76.29}\gain{+16.46}
& \textbf{96.16}\gain{+0.74}
& \textbf{56.41}\gain{+31.41}
& \textbf{57.01\gain{+8.59}} 
& \textbf{44.07\gain{+11.68}} 
& \textbf{80.31\gain{+7.66}} 
& \textbf{43.00\gain{+15.77}} 
& \textbf{43.59\gain{+20.91}} 
& \textbf{9.38\gain{+2.11}} \\
\bottomrule
\end{tabular}
}
\end{table*}

\section{Experiments}

\subsection{Experimental Setup}
To systematically evaluate our proposed method and compare it against baseline approaches under long-horizon clinical prediction settings, we conduct all experiments on the EHRSHOT benchmark~\cite{wornow2023ehrshot}. Unlike commonly used EHR dataset, such as MIMIC-III/IV~\cite{johnson2016mimic,johnson2023mimic}, which are largely restricted to ICU or emergency department settings, EHRSHOT captures longitudinal patient records across general hospital care. As a result, patient records in EHRSHOT often span multiple decades and contain thousands of clinical events. On average, EHRSHOT includes 2.3× more clinical events and 95.2× more encounters per patient than MIMIC-IV, making it substantially more challenging and better suited for evaluating long-horizon clinical reasoning. 

We select four representative prediction tasks from EHRSHOT: \textit{Long Length of Stay}, \textit{30-day Readmission}, \textit{Acute Myocardial Infarction (Acute MI)}, and \textit{Anemia}. The first two tasks fall under the \textit{Operational Outcomes} category, Anemia belongs to \textit{Anticipating Lab Test Results}, and Acute MI is categorized as \textit{Assignment of New Diagnoses}. The first three tasks are binary classification problems, while Anemia is a four-class classification task. Together, these tasks span diverse clinical objectives and temporal reasoning scenarios, enabling a comprehensive evaluation of long-horizon reasoning on structured EHR data. We report Accuracy, Macro-F1, and per-class F1 scores, as most clinical classification tasks are inherently imbalanced.

We conduct experiments using three LLMs that span both proprietary and open-source settings: \textit{GPT-5}, \textit{Claude-Opus-4.5}, and \textit{LLaMA-3.1-8B}. These models cover a broad range of architectures, training scales, model sizes, and accessibility levels, allowing us to assess the robustness of \method across different LLM backbones. Additional details of the experimental setup are provided in Appendix~\ref{ap:detailed_setup}.

\subsection{Baselines}
We compare \method against commonly used LLM-based baselines for EHR data, including direct generation and diverse RAG approaches.
\begin{itemize}[leftmargin=20pt, itemsep=0pt, labelsep=5pt, topsep=0pt]
    \item \textit{Direct Generation}~\cite{lin2025training}: The LLM directly predicts the clinical outcome from the patient EHR without retrieval. To satisfy context constraints, the EHR is truncated by retaining the most recent events up to the maximum context window, a common heuristic in prior LLM-based EHR studies. This baseline evaluates the performance of naive context truncation without evidence selection.
    \item \textit{RAG}~\cite{gao2023retrieval}: A vanilla retrieval approach that retrieves relevant EHR evidence and conditions the LLM on the retrieved context. This baseline reflects the most common single-pass RAG formulation used in recent clinical and non-clinical applications.
    \item \textit{Uniform RAG}~\cite{liu2024lost}: A retrieval baseline that uniformly samples EHR evidence without semantic prioritization. This baseline controls for context length and isolates the benefit of relevance-aware retrieval over random selection.
    \item \textit{Rule-based RAG}~\cite{liao2025ehr}: A heuristic retrieval approach that selects EHR evidence using predefined rules rather than learned relevance signals. In our implementation, events are ranked by occurrence frequency, and the top frequently occurring events are selected. This baseline reflects commonly adopted heuristic filtering strategies for handling long EHR sequences.
    \item \textit{ReAct RAG}~\cite{yao2022react}: A reasoning-and-acting framework that interleaves reasoning steps with retrieval actions. The LLM iteratively generates reasoning traces and retrieval queries to obtain additional EHR evidence, evaluating whether generic iterative retrieval improves long-horizon EHR reasoning without domain-specific design.
\end{itemize}

All baselines use the same underlying LLM backbone and context budget as \method to ensure a fair and controlled comparison.

Additionally, in Section~\ref{sec:ml_vs_llm}, we compare \method with classical ML baselines following the EHRSHOT benchmark to assess whether LLM-based reasoning offers benefits beyond classical EHR models:
\begin{itemize}[leftmargin=20pt, itemsep=0pt, labelsep=5pt, topsep=0pt]
    \item \textit{Count-based LR}~\cite{wornow2023ehrshot}: Logistic regression trained on count-based clinical features extracted from the EHR.
    \item \textit{CLMBR-based LR}~\cite{wornow2023ehrshot}: Logistic regression trained on patient representations produced by the pretrained \textit{CLMBR-T-Base} foundation model~\cite{wornow2023ehrshot,steinberg2021language}.
\end{itemize}

\begin{table*}[htbp]
\centering
\caption{Performance comparison of \method and other LLM-based baselines across different LLM backbones (\textit{GPT-5}, \textit{Claude-Opus-4.5}, and \textit{LLaMA-3.1-8B}) on the \textit{Long Length of Stay} task. We report Accuracy (\%), Macro-F1 (\%), and per-class F1 scores (\%). \textbf{Bold} indicates the best performance, and improvements over the strongest baseline are highlighted in \textcolor{langdarkgreen}{green}. The results demonstrate that the performance gains generalize across different LLM backbones.}
\label{tab:models}
\resizebox{\textwidth}{!}{
\begin{tabular}{lcccccccccccc}
\toprule
\multirow{2}{*}{\textbf{Method}} 
& \multicolumn{4}{c}{\textbf{GPT-5}} 
& \multicolumn{4}{c}{\textbf{Claude-Opus-4.5}} 
& \multicolumn{4}{c}{\textbf{LLaMA-3.1-8B}} \\
\cmidrule(lr){2-5} \cmidrule(lr){6-9} \cmidrule(lr){10-13}
& \textbf{Accuracy} 
& \textbf{Macro-F1} 
& \textbf{F1$_\text{Short}$} 
& \textbf{F1$_\text{Long}$}
& \textbf{Accuracy} 
& \textbf{Macro-F1} 
& \textbf{F1$_\text{Short}$} 
& \textbf{F1$_\text{Long}$}
& \textbf{Accuracy} 
& \textbf{Macro-F1} 
& \textbf{F1$_\text{Short}$} 
& \textbf{F1$_\text{Long}$} \\
\midrule
Direct Generation 
& 69.41 & 66.52 & 76.36 & 56.67
& 39.22 & 38.98 & 35.15 & 42.80
& 39.22 & 38.98 & 35.15 & 42.80 \\

RAG 
& 68.24 & 64.77 & 75.82 & 53.71
& 49.02 & 48.96 & 50.76 & 47.15
& 49.02 & 48.96 & 50.76 & 47.15 \\

Uniform RAG 
& 65.10 & 61.50 & 73.27 & 49.72
& 47.45 & 47.39 & 49.24 & 45.53
& 47.45 & 47.39 & 49.24 & 45.53 \\

Rule-based RAG 
& 63.53 & 56.58 & 73.95 & 39.22
& 45.10 & 45.03 & 46.97 & 43.09
& 45.10 & 45.03 & 46.97 & 43.09 \\

ReAct RAG 
& 65.10 & 60.85 & 73.75 & 47.95
& 47.84 & 47.76 & 45.71 & 49.81
& 47.84 & 47.76 & 45.71 & 49.81 \\

\midrule
\rowcolor{langlightgreen!50}\textbf{EHR-RAG (Ours)} 
& \textbf{74.12}\gain{+4.71} 
& \textbf{70.15}\gain{+3.63} 
& \textbf{81.03}\gain{+4.67} 
& \textbf{59.26}\gain{+2.59}
& \textbf{56.08}\gain{+7.06} 
& \textbf{55.58}\gain{+6.62} 
& \textbf{60.28}\gain{+9.52} 
& \textbf{50.88}\gain{+1.07}
& \textbf{56.08}\gain{+7.06} 
& \textbf{55.58}\gain{+6.62} 
& \textbf{60.28}\gain{+9.52} 
& \textbf{50.88}\gain{+1.07} \\
\bottomrule
\end{tabular}
}
\end{table*}

\begin{table}[ht]
\caption{Ablation results on the \textit{Long Length of Stay} and \textit{Anemia} tasks using \textit{GPT-5}. We report Accuracy (\%) and Macro-F1 (\%). \textcolor{langred}{Red} text indicates performance drops relative to \method.}
\label{tab:ablation}
\centering
\resizebox{\linewidth}{!}{
\begin{tabular}{lcccc}
\toprule
\multirow{2}{*}{\textbf{Method}} 
& \multicolumn{2}{c}{\textbf{Long Length of Stay}} 
& \multicolumn{2}{c}{\textbf{Anemia}} \\
\cmidrule(lr){2-3} \cmidrule(lr){4-5}
& Accuracy & Macro-F1
& Accuracy & Macro-F1 \\
\midrule
\rowcolor{langlightgreen!50}\textbf{EHR-RAG (Ours)} 
& \textbf{74.12} &  \textbf{70.15} &  \textbf{57.01} & \textbf{44.07} \\
\midrule
\quad w/o \textit{ETHER}
& 72.94\loss{-1.18} & 68.47\loss{-1.68} 
& 53.39\loss{-3.62} & 40.12\loss{-3.95} \\

\quad w/o \textit{AIR} 
& 72.16\loss{-1.96} & 68.19\loss{-1.96}
& 54.07\loss{-2.94} & 41.44\loss{-2.63} \\

\quad w/o \textit{DER} 
& 70.59\loss{-3.53} & 67.55\loss{-2.60} 
& 46.83\loss{-10.18} & 38.58\loss{-5.49} \\

Vanilla RAG 
& 68.24\loss{-5.88} & 64.77\loss{-5.38} 
& 42.08\loss{-14.93} & 20.44\loss{-23.63} \\

\bottomrule
\end{tabular}
}
\end{table}

\subsection{Main Results}

Tables~\ref{tab:main_los_readmission} and~\ref{tab:main_acutemi_anemia} present the main results on the four long-horizon clinical prediction tasks from EHRSHOT, spanning operational outcomes, diagnosis prediction, and laboratory abnormality assessment. 

\textbf{(i) Operational outcome prediction.}
As shown in Table~\ref{tab:main_los_readmission}, \method consistently outperforms all LLM-based baselines on both \textit{Long Length of Stay} and \textit{30-day Readmission}.
For length of stay prediction, \method achieves the highest accuracy (74.12) and macro-F1 (70.15), with substantial improvements over the strongest baseline across both short-stay and long-stay classes.
On the highly imbalanced readmission task, \method also yields notable improvements in macro-F1 (60.91).

\textbf{(ii) Clinical diagnosis and lab prediction.}
Table~\ref{tab:main_acutemi_anemia} presents results on \textit{Acute MI} and \textit{Anemia}.
For acute MI prediction, \method substantially improves macro-F1 (76.29) and F1 on the positive MI class (56.41), indicating more effective identification of clinically meaningful risk signals from long-horizon EHRs. On the multi-class anemia task, \method achieves the best overall accuracy (57.01) and macro-F1 (44.07), with consistent improvements across all severity levels, including the rare abnormal class.

\textbf{(iii) Baseline analysis.}
We observe that vanilla RAG can underperform direct generation in some cases, suggesting that retrieving events solely based on semantic similarity, without explicitly modeling recency and temporal context, may omit clinically critical recent evidence and lead to degraded performance. Uniform RAG exhibits unstable behavior, performing well in some cases but poorly in others, suggesting that while information across the patient history contains useful signals, indiscriminate retrieval lacks robustness. Rule-based RAG yields moderate improvements but introduces systematic bias by prioritizing high-frequency events, which can skew predictions toward dominant classes. ReAct RAG occasionally provides marginal gains over RAG by enabling iterative retrieval and broader evidence coverage; however, it does not consistently translate additional retrieval steps into reliable performance improvements.

\textbf{(iv) Overall comparison.}
Across all four tasks, \method consistently outperforms direct generation, vanilla RAG, and other retrieval baselines. These results highlight the importance of jointly modeling event-type structure, temporal dynamics, and evidence diversity through event- and time-aware retrieval, adaptive iterative refinement, and dual-path evidence retrieval and reasoning for accurate and reliable long-horizon clinical prediction from structured EHR data. Overall, these findings demonstrate that effective long-horizon EHR reasoning requires structured, temporally aware, and diversity-preserving retrieval mechanisms rather than generic or heuristic RAG strategies. Appendix~\ref{ap:case_study} presents a qualitative case study comparing \method with direct generation and vanilla RAG.

\subsection{Ablation Analysis of Key Components}

Table~\ref{tab:ablation} reports the ablation results on \textit{Long Length of Stay} and \textit{Anemia}, analyzing the contribution of each core component of \method. Red subscripts indicate the absolute performance drop relative to the full framework.

\textbf{(i) Overall impact of each component.}
Excluding \textit{Event- and Time-Aware Hybrid EHR Retrieval (ETHER)} leads to clear drops, particularly on \textit{Anemia}, highlighting the importance of preserving event-type distinctions and temporal structure when retrieving long-horizon clinical evidence. Removing \textit{Adaptive Iterative Retrieval (AIR)} results in uniform declines across all metrics, suggesting that iterative query refinement is critical for recovering sparse but task-relevant evidence distributed over time. Finally, ablating \textit{Dual-Path Evidence Retrieval Reasoning (DER)} causes the largest Macro-F1 degradation on \textit{Anemia}, underscoring the value of explicitly modeling factual and counterfactual retrieval and reasoning in complex and multi-class clinical prediction settings. Overall, the full framework achieves the best performance across all metrics, while each ablated variant exhibits systematic performance drops, demonstrating that all components of \method contribute meaningfully and synergistically to the overall gains.

\textbf{(ii) Comparison with vanilla RAG.}
Vanilla RAG performs substantially worse than all \method variants, with especially severe degradation on \textit{Anemia} (14.93 accuracy and 23.63 Macro-F1 points). This confirms that naive retrieval alone is insufficient for long-horizon structured EHR reasoning, and motivates the need for the proposed event-aware, iterative, and dual-path design.

\begin{figure*}[t]
    \centering
    \includegraphics[width=\textwidth]{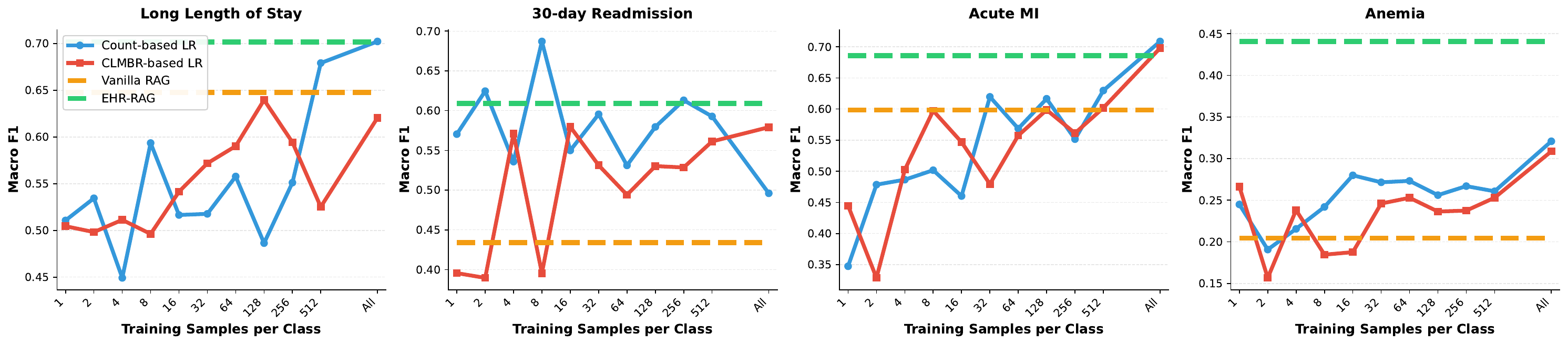}
    \caption{Macro-F1 performance comparison between \method, vanilla RAG, and traditional machine learning (ML) baselines under varying amounts of labeled training data across four tasks. The x-axis denotes the number of training samples per class, while dashed horizontal lines indicate the zero-shot performance of LLM-based methods. \method consistently matches or outperforms ML baselines, particularly in low-resource settings.}
    \label{fig:ml_vs_llm}
\end{figure*}

\subsection{Performance Across Different LLM Backbones}

Table~\ref{tab:models} compares the performance of different LLM backbones on the \textit{Long Length of Stay} task, including both proprietary and open-source models with varying capacities.

\textbf{(i) Consistent gains across models.}
\method yields consistent performance improvements across \textit{GPT-5}, \textit{Claude-Opus-4.5}, and \textit{LLaMA-3.1-8B}, despite substantial differences in model architecture and training scale. This consistency indicates that the proposed retrieval and reasoning framework generalizes across LLM backbones, rather than relying on the characteristics of any single model.

\textbf{(ii) Larger gains on smaller models.}
While \textit{GPT-5} achieves the highest absolute performance, the relative improvements introduced by \method are more pronounced for \textit{LLaMA-3.1-8B}. Compared to the strongest baseline, \method improves Macro-F1 by over 6 points on both models, suggesting that its design can partially mitigate the limitations of smaller model capacity.

\subsection{Comparison with Traditional ML Baselines}
\label{sec:ml_vs_llm}
Figure~\ref{fig:ml_vs_llm} compares the Macro-F1 performance of \method with ML baselines across varying levels of labeled data availability, ranging from few-shot to full-data regimes. The x-axis denotes the number of training samples per class. LLM-based methods are evaluated in a zero-shot setting and are therefore shown as horizontal reference lines.

\textbf{(i) Performance under low-data regimes.}
Across all tasks, \method outperforms ML baselines in most low-shot settings. When only a limited number of labeled samples per class are available (e.g., 1 to 16 shots), both count-based and CLMBR-based LR models experience substantial performance degradation. In contrast, \method maintains relatively stable performance by leveraging pretrained LLM knowledge together with retrieved EHR evidence, highlighting its strong data efficiency in low-resource settings.

\textbf{(ii) Comparison with representation-based ML models.}
As expected, CLMBR-based LR generally outperforms count-based features, particularly as the number of training samples increases, reflecting the advantage of pretrained foundation models in capturing general clinical representations. However, even in medium- and full-data settings, \method achieves comparable or superior Macro-F1 scores across all tasks. This suggests that retrieval-augmented LLM reasoning can effectively integrate heterogeneous and temporally distributed clinical evidence beyond what is captured by fixed patient representations.

\textbf{(iii) Robustness on long-horizon and imbalanced tasks.}
The advantages of \method are most pronounced on long-horizon and class-imbalanced tasks, such as \textit{Acute MI} and \textit{Anemia}. In these settings, ML baselines struggle to capture sparse yet clinically meaningful signals distributed over extended patient histories. In contrast, \method leverages strong generalization and zero-shot reasoning capabilities to surface and aggregate relevant evidence, leading to more robust performance.

Overall, these results demonstrate that \method not only outperforms traditional ML baselines in low-resource settings, but also remains competitive or superior even when large amounts of labeled data are available, underscoring its effectiveness as a general-purpose framework for long-horizon EHR prediction.

\section{Conclusion}
In this paper, we introduce \method, an enhanced retrieval-augmented framework designed to bridge long-horizon EHRs and LLMs. \method addresses key challenges posed by long-horizon EHR data, including context window limitations, temporal fragmentation, and incomplete evidence retrieval, through a combination of event- and time-aware hybrid EHR retrieval, adaptive iterative refinement, and dual-path evidence retrieval and reasoning. Extensive experiments demonstrate that \method consistently outperforms other LLM-based baselines across diverse long-horizon clinical prediction tasks. Overall, our results show that carefully designed retrieval and reasoning mechanisms can enable LLMs to effectively interpret long-horizon structured EHR data even under strict context constraints. We believe \method represents a practical step toward more reliable and data-efficient clinical prediction systems, and provides a foundation for future work on scalable, evidence-grounded reasoning over longitudinal healthcare data.




\clearpage

\section*{Impact Statement}
This work aims to advance large language model for reasoning over long-horizon electronic health records through retrieval-augmented generation. While our framework is developed using publicly accessible, de-identified data, we acknowledge that work involving clinical information raises considerations around privacy, fairness, and the appropriate use of AI-assisted decision support. Our method is designed as a research contribution rather than a deployed clinical system, and it does not make autonomous treatment recommendations. Instead, it seeks to improve technical robustness in temporal retrieval and reasoning, which may support future development of safer and more interpretable clinical AI tools.

\bibliography{references}
\bibliographystyle{icml2026}

\newpage
\appendix
\onecolumn

\section{Detailed Experimental Setup}
\label{ap:detailed_setup}

\textbf{Dataset Processing.}
We use the EHRSHOT benchmark~\cite{wornow2023ehrshot}, which is available in a MEDS-compatible format\footnote{\url{https://github.com/Medical-Event-Data-Standard}}. In the raw data, each clinical event is represented as a MEDS code. To make events interpretable to LLMs, we map these codes to natural-language descriptions by constructing a lightweight medical ontology from the Athena vocabulary\footnote{\url{https://athena.ohdsi.org/vocabulary/list}} and a CPT subset via UMLS\footnote{\url{https://uts.nlm.nih.gov/uts/}}. We evaluate four tasks, \textit{Long Length of Stay}, \textit{30-day Readmission}, \textit{Acute Myocardial Infarction}, and \textit{Anemia}, whose metadata are summarized in Appendix~\ref{ap:task_metadata}. For LLM inputs, we serialize each event using a unified template: \textit{\texttt{[time] type - description (value: value)}}; for example, \textit{\texttt{[2018-02-05 08:56:00] measurement - Blood Pressure (value: 120 mmHg)}}.

\textbf{LLM Generation Settings.}
We evaluate three LLMs: \textit{GPT-5} (\texttt{gpt-5-2025-08-07}), \textit{Claude-Opus-4.5} (\texttt{claude-opus-4-5-20251101}), and \textit{LLaMA-3.1-8B} (\texttt{Meta-Llama-3.1-8B-Instruct}). All models are deployed on the Azure AI platform, which complies with the PhysioNet Credentialed Data Use Agreement and responsible use guidelines for clinical data\footnote{\url{https://physionet.org/news/post/gpt-responsible-use}}. For all LLM-based methods, we set the generation temperature to 0 to ensure stable and deterministic outputs, except for \textit{GPT-5}, which requires a temperature of 1. To ensure computational efficiency and fair comparison across models, we fix the maximum context length to 128{,}000 tokens and keep all other generation hyperparameters at their default values. All prompt templates used in our experiments are provided in Appendix~\ref{ap:prompt_design}.

\textbf{ML Baseline Training Settings.}
The training setups for both \textit{Count-based LR} and \textit{CLMBR-based LR} strictly follow prior work~\cite{wornow2023ehrshot}, including model architectures and hyperparameter settings. For each task, we train separate models under varying data regimes with 1, 2, 4, 8, 16, 32, 64, 128, 256 shots per class, as well as using the full training set. The pretrained \textit{CLMBR-T-Base} foundation model~\cite{wornow2023ehrshot,steinberg2021language} is used to generate patient representations for the CLMBR-based LR baseline. All machine learning baselines are trained and evaluated on a single NVIDIA A800 GPU with 80GB of memory.

\textbf{Baseline Retrieval Settings.}
For \textit{Direct Generation}, we truncate each patient EHR by retaining the most recent 1{,}000 events. For \textit{Rule-based RAG}, we first rank all events by the frequency of their associated event codes within the patient history, and then select the top 1{,}000 most frequently occurring events as retrieved evidence. For all other RAG-based baselines, we adopt dense retrieval using \texttt{text-embedding-3-small} from Azure OpenAI as a lightweight text embedding model. Each patient EHR is segmented into chunks of 100 event rows with an overlap of 5 rows between adjacent chunks. For vanilla RAG methods, we retrieve the top 10 chunks per query, while keeping the total number of retrieved event rows comparable across methods. \textit{Uniform RAG} randomly samples chunks and concatenates them until the same event budget is reached, serving as a relevance-agnostic control. For \textit{ReAct RAG}, we retrieve the top 5 chunks per query over 3 iterative retrieval steps, accounting for potential duplicate retrievals across iterations. All retrieval pipelines are implemented using LangChain\footnote{\url{https://www.langchain.com}}, and cosine similarity between query and document embeddings is used as the retrieval scoring function.

\textbf{EHR-RAG Settings.}
For our proposed method, \method, we use the same dense text embedding model as the RAG baselines to ensure a fair comparison. The basic retrieval configuration follows \textit{ReAct RAG}: at each iteration, we retrieve the top-$K_{\text{final}}{=}5$ textual chunks per query over three iterative retrieval steps. In the U-shaped time-aware retrieval component, we set the semantic--temporal trade-off weight to $\alpha = 0.75$. The recent temporal window is defined as $\tau_{\text{recent}} = 180$ days, while the early-history window spans up to $\tau_{\text{early}} = 3{,}650$ days (approximately 10 years). For textual retrieval, an initial candidate pool of $K_{\text{cand}}{=}100$ chunks is retrieved before temporal re-ranking, from which the final evidence set $\mathcal{E}_{\text{text}}$ is selected. For numeric event retrieval, we adopt a two-stage indicator selection strategy with a coarse-grained top-$N_{\text{coarse}}{=}30$ followed by a fine-grained top-$N_{\text{fine}}{=}10$. For each selected indicator, we retain the $N_{\text{recent}}{=}5$ most recent measurements to construct the numeric evidence set $\mathcal{E}_{\text{num}}$. We observe that the overall performance of \method is not highly sensitive to these hyperparameters; they can be tuned intuitively within a reasonable range without significantly affecting results.


\clearpage
\section{Case Study}
\label{ap:case_study}

We present a detailed representative case study to illustrate the effectiveness of \method in clinical prediction, with a particular focus on its dual-path reasoning capability for balancing contradictory clinical evidence. This case demonstrates how \method avoids false alarms by systematically integrating both risk factors and stabilizing clinical signals, leading to more calibrated and reliable predictions. These results are generated using \textit{GPT-5} on the \textit{Long Length of Stay} task.

\textbf{Ethical and privacy note.} This case study is based on de-identified EHRSHOT data and does not include any personally identifiable information. It is intended solely to qualitatively illustrate retrieval and reasoning behavior, rather than to provide clinical guidance or diagnostic recommendations.

\subsection{Clinical Scenario}

\noindent\textbf{Patient:} Female, 56 years old; BMI 43.0 \\
\textbf{Admission Date:} 2014-12-05 20{:}00 (postoperative admission after intracranial aneurysm clipping) \\
\textbf{Prediction Time:} 2014-12-05 23{:}59 \\
\textbf{Ground Truth Outcome:} LOS = 1 day (discharged 2014-12-06) \\
\textbf{Task:} Predict whether long LOS ($\geq$ 7 days)

\medskip
\noindent\textbf{Clinical Context.}
Initially admitted on 2014-11-27 for subarachnoid hemorrhage (SAH) due to ruptured intracranial aneurysm; discharged 2014-11-28 after conservative management. Readmitted on 2014-12-04 for elective intracranial aneurysm clipping. The microsurgical procedure was completed successfully, followed by routine postoperative monitoring in the neurological intensive care unit.

\medskip
\noindent\textbf{Baseline Comorbidities.}
Type 2 diabetes mellitus / Hypertension / Hyperlipidemia / Class III obesity (BMI 43.0) / Tobacco use

\medskip
\noindent\textbf{Model Predictions and Outcomes.}
\begin{itemize}[leftmargin=*, itemsep=2pt, labelsep=5pt, topsep=2pt]
    \item \textbf{\method:} Prediction = 0 \;(\textcolor{langdarkgreen}{\textbf{Correct}}). \;
    \textit{Rationale:} Balances surgical risk factors with stabilizing postoperative signals, including rapid neurologic recovery and early clinical stability.
    \item \textbf{Direct Generation:} Prediction = 1 \;(\textcolor{langred}{\textbf{Incorrect}}). \;
    \textit{Rationale:} Overemphasizes surgical complexity and baseline comorbidities, overestimating prolonged hospitalization risk.
    \item \textbf{Vanilla RAG:} Prediction = 1 \;(\textcolor{langred}{\textbf{Incorrect}}). \;
    \textit{Rationale:} Retrieved evidence is dominated by high-risk features without incorporating countervailing stabilization indicators, yielding a biased risk assessment.
\end{itemize}

\subsection{Event- and Time-aware Hybrid EHR Retrieval in \method}

\noindent The \textit{Event- and Time-aware Hybrid EHR Retrieval} (ETHER) module first performs knowledge-guided coarse retrieval to collect candidate clinical indicators, then applies LLM-based reranking to select task-relevant indicators for downstream reasoning.

\subsubsection{Coarse Retrieval}
\noindent Using medical priors for postoperative neurologic and respiratory monitoring, ETHER retrieves candidates spanning: neurologic status (e.g., GCS), respiratory parameters (e.g., PEEP, FiO$_2$, airway pressure), arterial blood gases (pH, PaCO$_2$, PaO$_2$), hemodynamics (blood pressure), and coagulation markers (INR, aPTT), as well as additional candidates including vasopressor use, ventilation settings, pain, and sedation metrics.

\subsubsection{LLM-based Indicator Reranking}
\noindent The LLM reranks these candidates and selects the top-10 indicators:
\begin{enumerate}[leftmargin=*, itemsep=1pt, topsep=2pt]
    \item Glasgow Coma Scale (GCS)
    \item Arterial blood gas measures (pH, PaCO$_2$, PaO$_2$)
    \item Fraction of inspired oxygen (FiO$_2$)
    \item Mean airway pressure and PEEP
    \item Central venous pressure (CVP)
    \item Diastolic blood pressure (DBP)
    \item International Normalized Ratio (INR)
\end{enumerate}

\subsubsection{Retrieved Clinical Evidence}

\noindent\textbf{Hemodynamics and Vital Signs.}
Stable SBP 98--130 mmHg, DBP 57--64 mmHg, and HR 81--89 bpm throughout the postoperative period; no vasopressor support required. Mean arterial pressure is consistently maintained within a normal range.

\noindent\textbf{Neurologic Status.}
Rapid neurologic recovery: transient post-anesthetic decline (GCS 14) improves to full responsiveness (GCS 15). Pain decreases from 6/10 to 0/10 within two hours postoperatively; spontaneous respiration is preserved.

\noindent\textbf{Respiratory and Blood Gas Trends.}
Expected postoperative disturbances with rapid normalization: mild acidosis (pH 7.24) and hypercapnia (PaCO$_2$ 52 mmHg) improve within two hours (pH 7.34, PaCO$_2$ 46.5 mmHg). Oxygenation remains adequate; FiO$_2$ is weaned from 60\% to 50\%, indicating improving respiratory status with minimal ventilatory support.

\noindent\textbf{Coagulation and Metabolic Status.}
Normal coagulation (INR 1.1; aPTT 21.9 s); preserved renal function (creatinine 0.9 mg/dL); electrolytes stable. Transient hyperglycemia (consistent with postoperative stress) is observed.

\noindent\textbf{Monitoring and Support.}
Standard postoperative monitoring is in place, including arterial and central venous lines and temporary airway protection, without evidence of escalating support requirements.

\medskip
\noindent\textbf{Summary.}
Overall, the retrieved evidence indicates rapid postoperative stabilization across neurologic, respiratory, and hemodynamic domains, providing strong signals against prolonged hospitalization despite high baseline risk factors.

\subsection{Adaptive Information Retrieval (AIR) in \method}

\noindent AIR refines retrieval by adaptively generating positive and negative queries to gather evidence supporting both hypotheses.

\subsubsection{Positive Reasoning Path (Evidence for Long Stay)}

\noindent\textbf{Iteration 0:} Query = ``Cerebral vasospasm following SAH''
\begin{itemize}[leftmargin=*, itemsep=1pt, topsep=2pt]
    \item Retrieved: Vasospasm monitoring protocols, CBF measurements, transcranial doppler assessments (5 documents).
    \item Rationale: Vasospasm is a major SAH complication requiring prolonged ICU stay; early monitoring is critical.
\end{itemize}

\noindent\textbf{Iteration 1:} Query = ``Hydrocephalus following intracranial hemorrhage''
\begin{itemize}[leftmargin=*, itemsep=1pt, topsep=2pt]
    \item Retrieved: Ventriculomegaly assessments, intracranial pressure trends, EVD indications (6 documents).
    \item Rationale: SAH frequently causes hydrocephalus; EVD insertion commonly prolongs ICU stay.
\end{itemize}

\noindent\textbf{Iteration 2:} Query = ``External ventricular drain management post-aneurysm''
\begin{itemize}[leftmargin=*, itemsep=1pt, topsep=2pt]
    \item Retrieved: EVD care protocols, CSF characteristics, EVD removal criteria, duration of drainage (3 documents).
    \item Rationale: EVD presence strongly predicts extended ICU stay (typically 5--7+ days).
\end{itemize}

\subsubsection{Negative Reasoning Path (Evidence for Short Stay)}

\noindent\textbf{Iteration 0:} Query = ``Discharge to home post-neurosurgery without complications''
\begin{itemize}[leftmargin=*, itemsep=1pt, topsep=2pt]
    \item Retrieved: Home discharge orders, absence of postoperative complications, successful emergence from anesthesia (4 documents).
    \item Rationale: Home discharge after neurosurgery indicates rapid, uncomplicated recovery.
\end{itemize}

\noindent\textbf{Iteration 1:} Query = ``Social work assessment and discharge planning after ICU''
\begin{itemize}[leftmargin=*, itemsep=1pt, topsep=2pt]
    \item Retrieved: Discharge readiness criteria, caregiver assessments, skilled nursing vs.\ home discharge (5 documents).
    \item Rationale: Early discharge planning and lack of complications facilitate short stay.
\end{itemize}

\noindent\textbf{Iteration 2:} Query = ``Physical therapy: independent ambulation post-operative''
\begin{itemize}[leftmargin=*, itemsep=1pt, topsep=2pt]
    \item Retrieved: PT clearance for discharge, functional independence measures, ICU mobility protocols (3 documents).
    \item Rationale: Independent ambulation signals discharge readiness.
\end{itemize}

\subsection{Dual Evidence Retrieval and Reasoning (DER) in \method}

\noindent DER evaluates evidence for long- vs.\ short-stay hypotheses independently and then fuses contradictory evidence.

\subsubsection{Positive Reasoning (Supporting Long Stay $\geq$ 7 Days)}

\noindent\textbf{Severe Diagnosis and Procedure Complexity:}
\begin{itemize}[leftmargin=*, itemsep=1pt, topsep=2pt]
    \item Intracranial aneurysm clipping is a high-risk, complex microsurgical procedure.
    \item Requires precise dissection of carotid circulation and careful handling of fragile vessel.
    \item Typical post-op course involves 5--7 day minimum ICU stay for neuro monitoring.
    \item SAH increases risk of post-operative complications.
\end{itemize}

\noindent\textbf{High-Intensity ICU Care Requirements:}
\begin{itemize}[leftmargin=*, itemsep=1pt, topsep=2pt]
    \item Mechanical ventilation for airway protection post-op.
    \item Continuous neuro monitoring (GCS, pupil reactivity, motor/sensory).
    \item Arterial line for beat-to-beat BP monitoring (neurosurgery standard).
    \item Central venous line for volume assessment and medication administration.
    \item Thermoregulation management (hypothermia induction protocol common in SAH).
    \item Vasoactive medications (nitroprusside documented) for hemodynamic optimization.
\end{itemize}

\noindent\textbf{Acute Physiologic Derangements:}
\begin{itemize}[leftmargin=*, itemsep=1pt, topsep=2pt]
    \item Early metabolic acidosis (pH 7.24) indicating post-operative stress.
    \item Hypercapnia (PaCO$_2$ 52 mmHg) suggesting ventilation/perfusion mismatch.
    \item Stress hyperglycemia (glucose 235 mg/dL) indicating systemic metabolic stress.
    \item These derangements typically persist for 3--5 days, prolonging ICU stay.
\end{itemize}

\noindent\textbf{High-Risk Comorbidity Profile:}
\begin{itemize}[leftmargin=*, itemsep=1pt, topsep=2pt]
    \item Type 2 diabetes increases post-operative infection risk and impairs healing.
    \item Hypertension complicates post-op BP management in neurosurgery.
    \item Obesity increases anesthetic risk and mobility limitations.
    \item Tobacco use associated with respiratory complications.
    \item Combination of comorbidities predicts longer recovery trajectory.
\end{itemize}

\noindent\textbf{Rehabilitation and Disposition Planning:}
\begin{itemize}[leftmargin=*, itemsep=1pt, topsep=2pt]
    \item Documented rehabilitation facility admission suggests anticipated prolonged stay.
    \item Comorbidities and mobility limitations often necessitate post-acute care facility.
    \item Home discharge typically requires minimal comorbidities and full functional independence.
\end{itemize}

\subsubsection{Negative Reasoning (Supporting Short Stay $<$ 7 Days)}

\noindent\textbf{Rapid Anesthetic Emergence and Neurologic Recovery:}
\begin{itemize}[leftmargin=*, itemsep=1pt, topsep=2pt]
    \item By 20{:}42 (42 minutes post-op): MAC 0--0.1 indicates patient nearly fully awake.
    \item By 21{:}15 (75 minutes post-op): GCS 15 with full consciousness and appropriate responses.
    \item No requirement for prolonged sedation or neuromuscular blockade.
    \item Rapid emergence suggests excellent anesthetic recovery and minimal post-op CNS depression.
\end{itemize}

\noindent\textbf{Stable Hemodynamics Without Vasopressor Dependence:}
\begin{itemize}[leftmargin=*, itemsep=1pt, topsep=2pt]
    \item SBP 98--130 mmHg without vasopressor infusions; mean BP 70--90s sufficient for cerebral perfusion.
    \item No hypotensive episodes requiring intervention.
    \item No documented inotropic support (dopamine, epinephrine).
    \item Hemodynamic stability by 1--2 hours post-op is a strong negative predictor.
\end{itemize}

\noindent\textbf{Preserved Neurologic Integrity:}
\begin{itemize}[leftmargin=*, itemsep=1pt, topsep=2pt]
    \item GCS 15 at 21{:}15 demonstrates full consciousness.
    \item Temporary GCS 14 at 22{:}00 is mild and transient (likely opioid-related).
    \item No focal neurologic deficits; no stroke or intraoperative ischemic complications.
    \item Neurologic preservation suggests successful repair without complications.
\end{itemize}

\noindent\textbf{Excellent Pain Control and Comfort:}
\begin{itemize}[leftmargin=*, itemsep=1pt, topsep=2pt]
    \item Pain reduced from 6/10 to 0/10 within 45 minutes.
    \item Adequate analgesia supports early mobilization and reduced ICU stay.
\end{itemize}

\noindent\textbf{Laboratory Normalization and No Coagulopathy:}
\begin{itemize}[leftmargin=*, itemsep=1pt, topsep=2pt]
    \item Creatinine 0.9 mg/dL (normal renal function); electrolytes within normal limits.
    \item INR 1.1 and aPTT 21.9 indicate normal coagulation; no transfusions documented.
    \item Lack of metabolic/coagulation complications reduces ICU stay duration.
\end{itemize}

\noindent\textbf{Physiologic Improvement and ABG Trend:}
\begin{itemize}[leftmargin=*, itemsep=1pt, topsep=2pt]
    \item pH improves from 7.24 to 7.34 within 2 hours.
    \item PaCO$_2$ decreases from 52 to 46.5 mmHg; base deficit improves from 5 to 1 mmol/L.
    \item Rapid recovery suggests minimal complications and improving discharge readiness.
\end{itemize}

\noindent\textbf{Absence of Documented Post-operative Complications:}
\begin{itemize}[leftmargin=*, itemsep=1pt, topsep=2pt]
    \item No rebleeding, hematoma expansion, new stroke/ischemia, ventilator failure, or respiratory distress.
    \item No infection/fever/sepsis; no escalation to vasopressors/ECMO.
\end{itemize}

\noindent\textbf{Prior Rapid Recovery Pattern:}
\begin{itemize}[leftmargin=*, itemsep=1pt, topsep=2pt]
    \item Prior SAH admission (11/27) discharged within 1 day (11/28), indicating rapid recovery.
    \item Readmission on 12/04 for planned elective repair (not emergency).
    \item Planned procedures typically have better outcomes than emergency cases.
    \item Patient's prior trajectory supports expectation of rapid current recovery.
\end{itemize}

\subsubsection{Evidence Fusion}

\noindent The DER fusion module synthesizes both positive and negative evidence:
\begin{quote}\small
\textit{``While the procedure itself is complex and carries inherent risks typical of intracranial aneurysm repair, this particular patient demonstrated exceptionally rapid post-operative recovery. Within 2 hours of ICU admission, she was fully conscious (GCS 15), hemodynamically stable without vasopressor support, and showing rapid physiologic improvement (ABG normalization, pain control achieved).\\
The presence of comorbidities (diabetes, obesity, HTN) and historical SAH are concerning for prolonged stay. However, the complete absence of any documented post-operative complication, combined with the patient's prior history of rapid discharge after SAH and the elective (not emergency) nature of this procedure suggest excellent surgical repair with minimal morbidity.\\
By prediction time (23{:}59, 4 hours post-op), all indicators point to an uncomplicated trajectory: no vasospasm signs, no hydrocephalus indicators, no coagulopathy, normal renal function, and preserved neurologic status. This contrasts with the typical 5--7 day post-op course.\\
\textbf{Conclusion:} The balance of evidence strongly favors shorter length of stay ($<$7 days). The rapid recovery trajectory and absence of complications outweigh theoretical risks from procedure complexity and comorbidities.''}
\end{quote}

\noindent\textbf{Final Prediction:} 0 (Length of Stay $<$ 7 days) \hfill
\textbf{Confidence:} High \\
\textbf{Ground Truth:} Discharged 2014-12-06 (1 day post-op) \hfill
\textbf{Outcome:} \textcolor{langdarkgreen}{\textbf{\checkmark Correct}}

\subsection{Failure Analysis: Why Direct Generation and Vanilla RAG Failed}

\subsubsection{Direct Generation Failure}

\noindent\textbf{Reasoning Pattern:}
\begin{quote}\small
\textit{``Complex intracranial aneurysm clipping surgery in a patient with multiple comorbidities (diabetes, obesity, hypertension) typically requires a lengthy ICU stay for monitoring of post-operative complications such as vasospasm and hydrocephalus. The patient's initial presentation with SAH suggests high-risk recovery. Expected LOS $\geq$ 7 days.''}
\end{quote}

\noindent\textbf{Failure Mode -- Confirmation Bias:}
\begin{itemize}[leftmargin=*, itemsep=1pt, topsep=2pt]
    \item \textbf{Received Data:} Accessed identical clinical data (GCS, ABGs, vital signs, etc.).
    \item \textbf{Reasoning Error:} Anchored on surgical complexity and comorbidities, underweighting recovery trajectory.
    \item \textbf{Bias:} Overweighted worst-case complications (vasospasm, hydrocephalus) without evidence.
    \item \textbf{Missed Signal:} Failed to treat rapid neurologic recovery (GCS 15 within hours) as countervailing evidence.
    \item \textbf{Static Thinking:} Evaluated a single timepoint rather than dynamic improvement.
\end{itemize}

\subsubsection{Vanilla RAG Failure}

\noindent\textbf{Reasoning Pattern:}
\begin{quote}\small
\textit{``Retrieval augmented generation returned articles on post-operative complications of intracranial aneurysm surgery, typical ICU stay durations (5--7 days), and management of vasospasm and hydrocephalus. Based on retrieved context, predicted LOS $\geq$ 7 days.''}
\end{quote}

\noindent\textbf{Failure Mode -- Lack of Systematic Evaluation:}
\begin{itemize}[leftmargin=*, itemsep=1pt, topsep=2pt]
    \item \textbf{Retrieval Bias:} Query biased toward complications and prolonged monitoring.
    \item \textbf{Missing Dual Perspective:} Retrieved risk-confirming documents without systematic search for rapid recovery evidence.
    \item \textbf{No Negative Evidence:} No structured mechanism to identify/weight evidence against high-risk conclusion.
    \item \textbf{Static Aggregation:} Treated typical 5--7 day LOS as a rule, ignoring patient-specific trajectory.
    \item \textbf{Confirmation Loop:} Retrieval reinforced initial impression without critical reappraisal.
\end{itemize}

\subsubsection{EHR-RAG Success}

\noindent\textbf{Key Advantages:}
\begin{itemize}[leftmargin=*, itemsep=1pt, topsep=2pt]
    \item \textbf{Systematic Dual Evaluation:} Explicitly considers both positive and negative evidence.
    \item \textbf{Balanced Retrieval:} AIR issues positive queries (vasospasm, hydrocephalus) and negative queries (recovery, discharge planning).
    \item \textbf{Trajectory Analysis:} DER emphasizes dynamic trends (e.g., ABG improvement) rather than static risk factors.
    \item \textbf{Evidence Fusion:} Appropriately weights rapid improvement against theoretical risks.
    \item \textbf{False-alarm Reduction:} Prevents overconfident high-risk classification via structured dual reasoning.
\end{itemize}

\subsection{Clinical Significance and Learning Points}

\noindent\textbf{Key Insight.}
Complex surgery does not automatically imply prolonged stay:
\begin{enumerate}[leftmargin=*, itemsep=1pt, topsep=2pt]
    \item \textbf{Procedure complexity $\neq$ complication inevitability:} Skilled repair can avoid major post-op complications.
    \item \textbf{Rapid recovery in high-risk patients:} Despite obesity/diabetes/HTN, full neurologic recovery occurs within hours.
    \item \textbf{Trajectory over static risk:} Improving vitals/labs/neuro status within 2 hours is a strong predictor of short stay.
    \item \textbf{No early complications:} Absence of complications by 4 hours post-op increases short-stay likelihood.
\end{enumerate}

\noindent\textbf{Clinical Impact of Accurate Prediction:}
\begin{itemize}[leftmargin=*, itemsep=1pt, topsep=2pt]
    \item \textbf{ICU resource allocation:} Enables earlier step-down and frees beds for critical patients.
    \item \textbf{Discharge planning:} Supports planning for rapid discharge vs.\ prolonged facility stay.
    \item \textbf{Patient/family expectations:} Reduces unnecessary anxiety about prolonged recovery.
    \item \textbf{Cost reduction:} Avoids unnecessary ICU days (at \textasciitilde\$5000/day).
    \item \textbf{Mobility/infection prevention:} Earlier step-down supports mobilization and reduces ICU-related risks.
\end{itemize}

\subsection{Summary}
The structured dual reasoning of EHR-RAG mitigates false-alarm bias by preventing worst-case anchoring without proportional evidence. By enforcing balanced evaluation of contradictory evidence, EHR-RAG maintains sensitivity to high-risk cases while reducing excessive false positives.

\clearpage
\section{Metadata for Clinical Prediction Tasks}
\label{ap:task_metadata}

Table~\ref{tab:task_metadata} presents the metadata of the EHRSHOT clinical prediction tasks used in our experiments, including core task definitions, basic task queries, and task-specific instructions. This metadata is provided to the LLMs under a unified prompt framework to support clinical prediction across different tasks.

\begin{table*}[htbp]
\caption{Metadata of EHRSHOT clinical prediction tasks used in our experiments. Each column corresponds to a task, and each row reports a specific metadata field.}
\label{tab:task_metadata}
\centering
\resizebox{\textwidth}{!}{
\renewcommand{\arraystretch}{1.5}
\begin{tabular}{p{3.5cm} p{4.3cm} p{4.3cm} p{4.3cm} p{4.3cm}}
\toprule
\textbf{Field} & \textbf{guo\_los} & \textbf{guo\_readmission} & \textbf{new\_acutemi} & \textbf{lab\_anemia} \\
\midrule

Task name &
Long Length of Stay &
30-day Readmission &
Acute Myocardial Infarction &
Anemia \\

Category &
Operational Outcomes &
Operational Outcomes &
Assignment of New Diagnoses &
Anticipating Lab Test Results \\

Description &
Predict whether a patient will have a long hospital stay ($\ge$7 days) based on their EHR data. &
Predict whether a patient will be readmitted within 30 days after hospital discharge based on their EHR data. &
Predict whether a patient will receive a new acute myocardial infarction diagnosis within 1 year after discharge based on their EHR data. &
Predict the severity category of the next anemia-related laboratory result based on the patient's prior EHR data. \\

Factual query &
clinical factors and events associated with prolonged hospital stay &
clinical factors and events associated with 30-day hospital readmission &
clinical risk factors and events associated with acute myocardial infarction &
clinical factors and events relevant to predicting anemia severity \\

Counterfactual query &
clinical factors and events associated with short hospital stay &
clinical factors and events associated with no readmission &
clinical risk factors and events indicating absence of acute myocardial infarction &
clinical factors and events indicating no anemia \\

Label type &
binary &
binary &
binary &
multiclass\_4 \\

Label values &
$\{0,1\}$ &
$\{0,1\}$ &
$\{0,1\}$ &
$\{0,1,2,3\}$ \\

Label description &
$\{0,1\}$ aka \{<7 days, $\ge$7 days\} &
$\{0,1\}$ aka \{no readmission, readmission\} &
$\{0,1\}$ aka \{no diagnosis, diagnosis\} &
$\{0,1,2,3\}$ aka \{low, medium, high, abnormal\} \\

Task-specific instructions &
No instructions provided. &
30-day readmission is uncommon. Default to 0 unless there is clear, strong, patient-specific evidence; do not predict 1 from vague risk factors. &
Be sensitive to positives: if there is any reasonable, patient-specific evidence suggesting acute MI, lean toward 1; if uncertain, prefer 1. &
Choose among \{0,1,2,3\} with calibrated preference for mild-to-moderate (1 or 2) when uncertain; use 0 only with strong evidence of no anemia; use 3 only with clear severe/abnormal anemia evidence. \\

\bottomrule
\end{tabular}
}
\end{table*}

\clearpage
\section{Prompt Design}
\label{ap:prompt_design}

\begin{figure*}[htbp]
    \centering
    \includegraphics[width=0.8\textwidth]{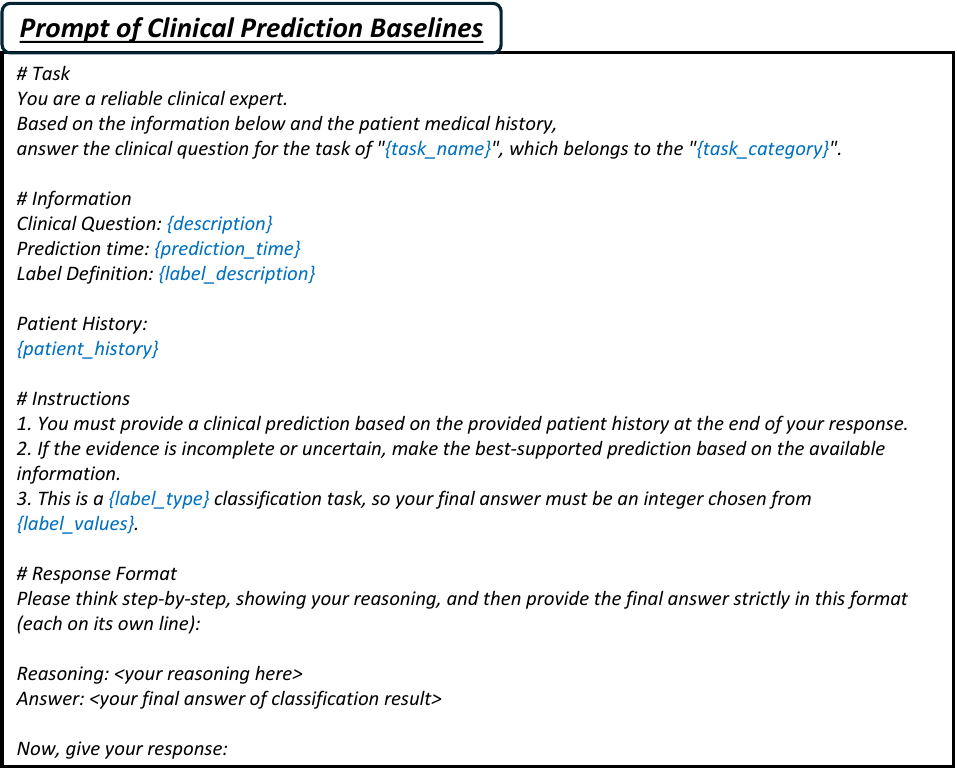}
    \caption{Prompt template used for LLM-based clinical prediction baselines. Blue text indicates input variables.}
    \label{prompt:baselines}
\end{figure*}

\begin{figure*}[htbp]
    \centering
    \includegraphics[width=0.8\textwidth]{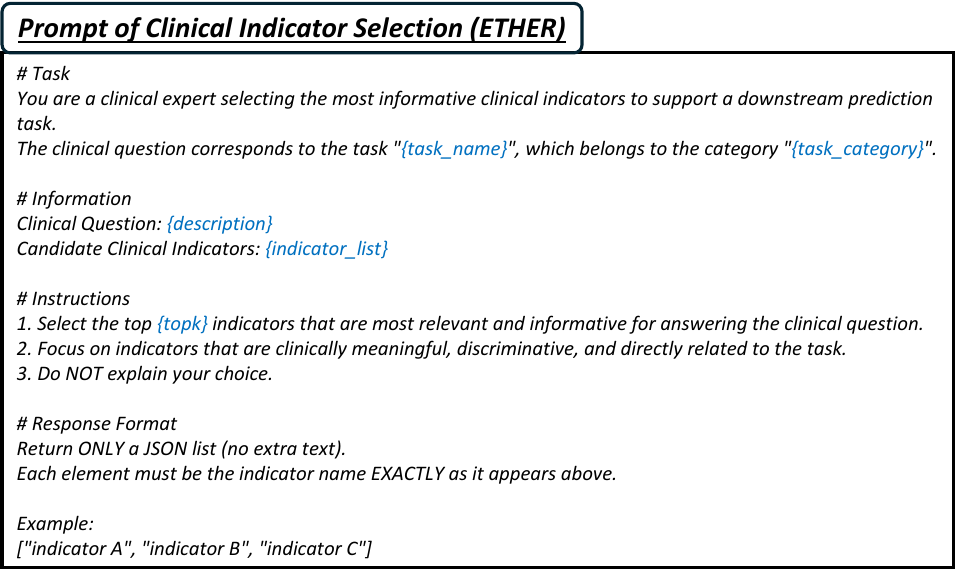}
    \caption{Prompt template used for clinical indicator selection in the \textit{Event- and Time-Aware Hybrid EHR Retrieval} component of \method. Blue text denotes input variables.}
    \label{prompt:select}
\end{figure*}

\begin{figure*}[htbp]
    \centering
    \includegraphics[width=0.8\textwidth]{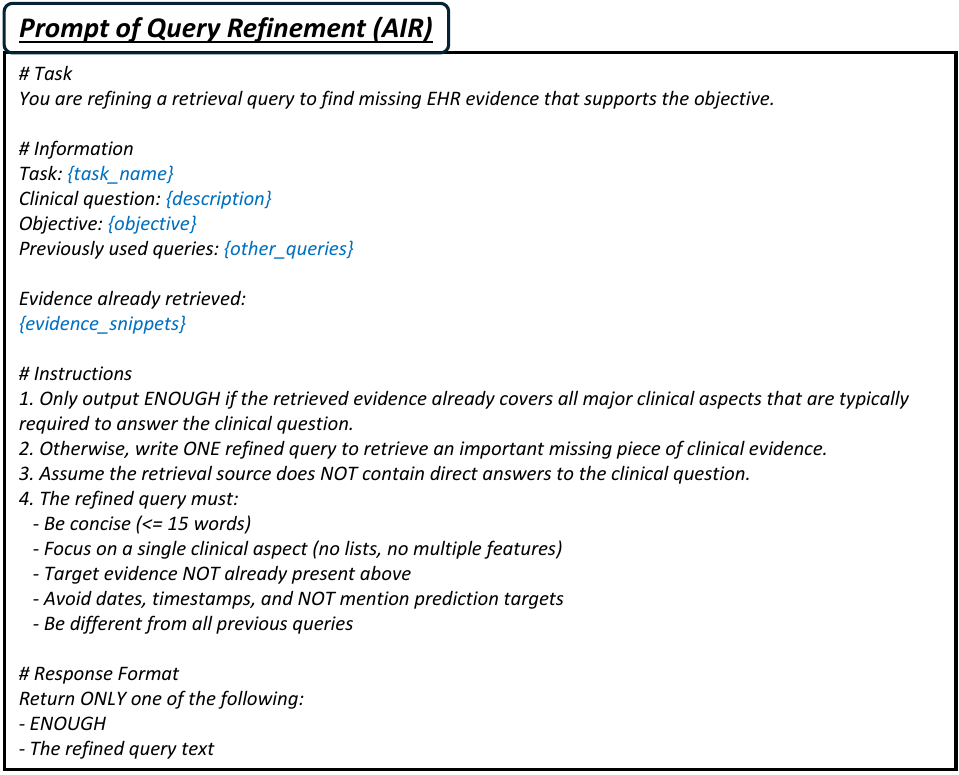}
    \caption{Prompt template used for query refinement in the \textit{Adaptive Iterative Retrieval} component of \method. Blue text denotes input variables.}
    \label{prompt:query_refine}
\end{figure*}

\begin{figure*}[htbp]
    \centering
    \includegraphics[width=0.8\textwidth]{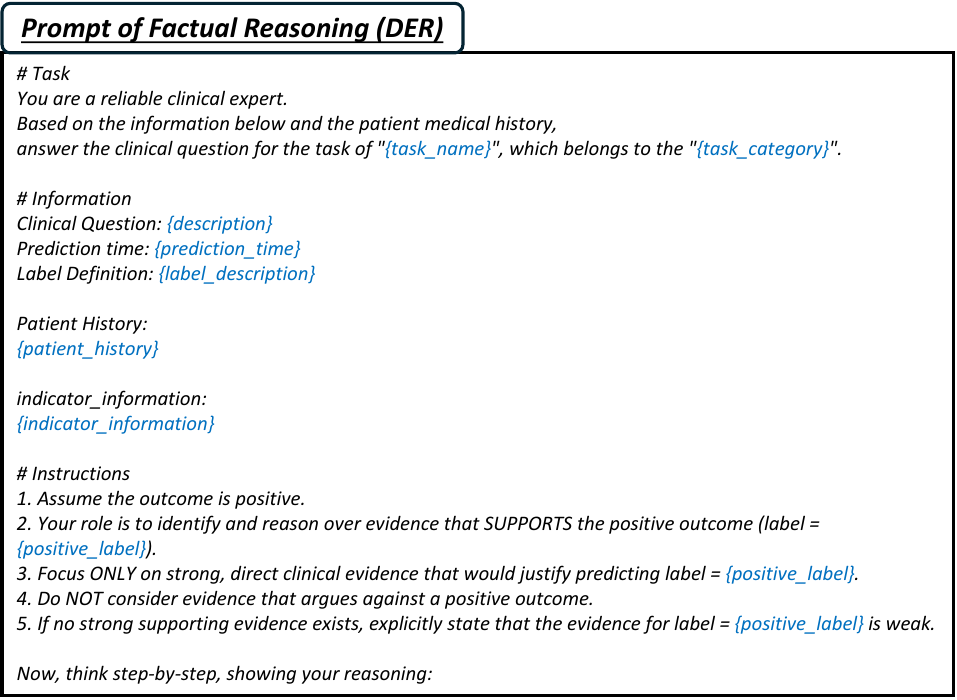}
    \caption{Prompt template used for factual reasoning in the \textit{Dual-Path Evidence Retrieval and Reasoning} component of \method. Blue text denotes input variables.}
    \label{prompt:query_factual}
\end{figure*}

\begin{figure*}[htbp]
    \centering
    \includegraphics[width=0.8\textwidth]{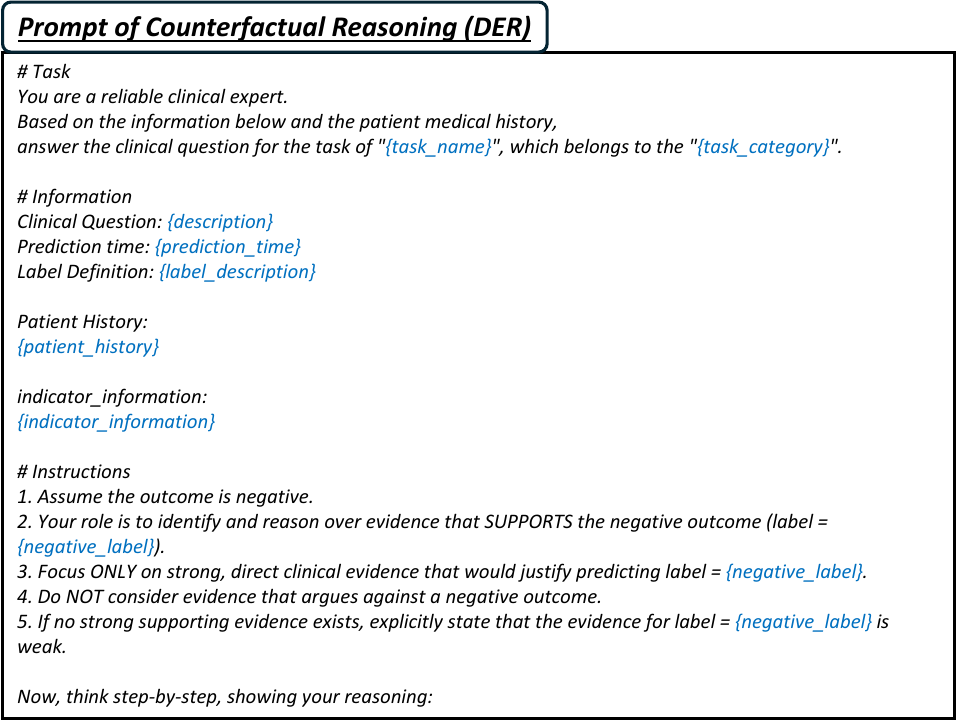}
    \caption{Prompt template used for counterfactual reasoning in the \textit{Dual-Path Evidence Retrieval and Reasoning} component of \method. Blue text denotes input variables.}
    \label{prompt:query_counterfactual}
\end{figure*}

\begin{figure*}[htbp]
    \centering
    \includegraphics[width=0.8\textwidth]{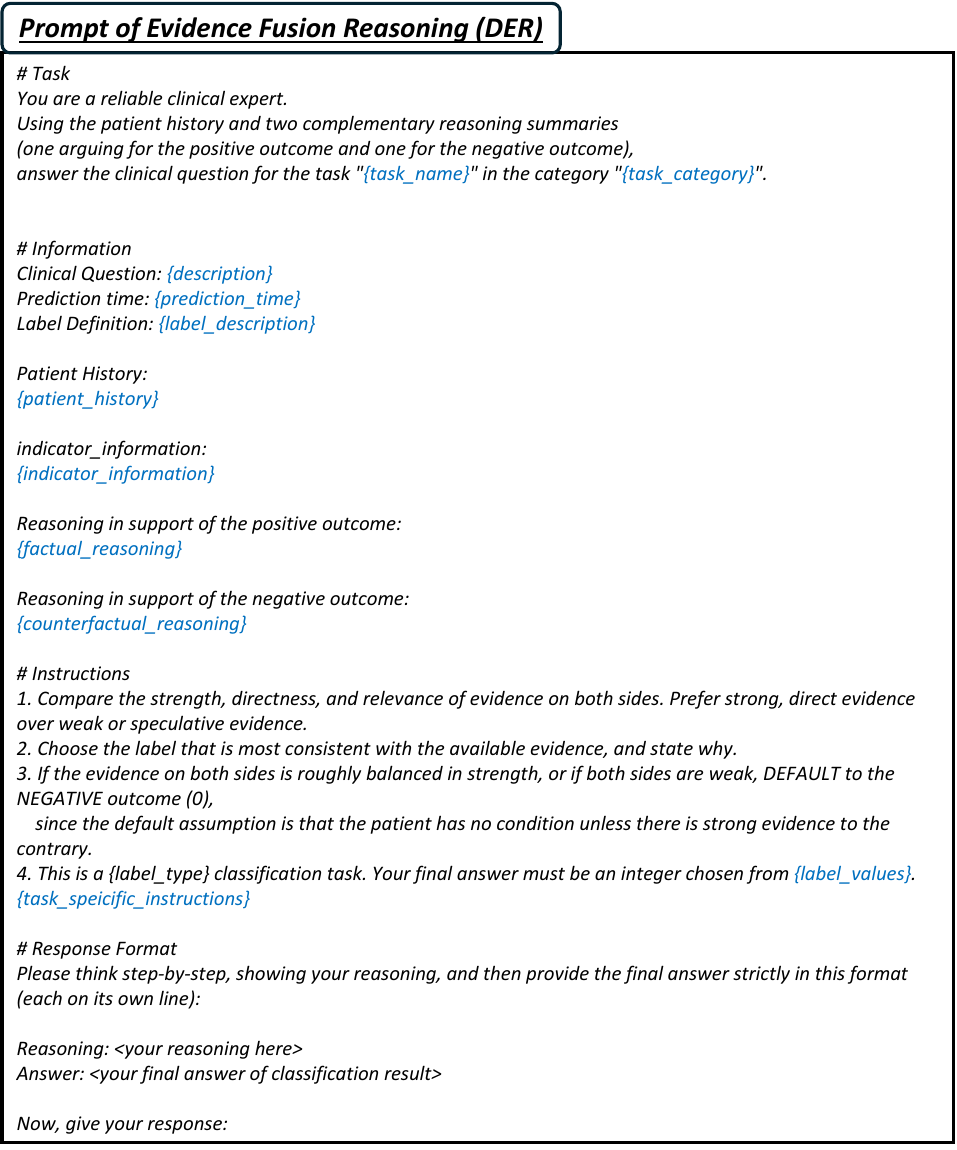}
    \caption{Prompt template used for evidence fusion reasoning in the \textit{Dual-Path Evidence Retrieval and Reasoning} component of \method. Blue text denotes input variables.}
    \label{prompt:query_fuse}
\end{figure*}

Figures~\ref{prompt:baselines} present the prompt template used by the LLM-based clinical prediction baselines.
Figures~\ref{prompt:select} and \ref{prompt:query_refine} illustrate the prompts employed in the \textit{Event- and Time-Aware Hybrid EHR Retrieval (ETHER)}  and \textit{Adaptive Iterative Retrieval (AIR)} components of \method, respectively. Figures~\ref{prompt:query_factual}, \ref{prompt:query_counterfactual}, and \ref{prompt:query_fuse} show the prompt templates used for factual reasoning, counterfactual reasoning, and evidence fusion in the \textit{Dual-Path Evidence Retrieval and Reasoning (DER)} component.

\end{document}